\documentclass[runningheads]{llncs}
\usepackage[T1]{fontenc}
\usepackage{graphicx}
\usepackage{booktabs}
\usepackage[misc]{ifsym}

\usepackage{amsmath,amssymb}
\usepackage{algorithm}
\usepackage{algpseudocode}
\usepackage{tikz}
\usepackage{hyperref}
\usepackage{url}

\begin{document}

\title{Implementing Causal Perception: \\Competing SCMs and Situated Fairness}
\titlerunning{Implementing Causal Perception}

\author{Jos\'e M. \'Alvarez\orcidID{0000-0001-9412-9013}\thanks{Forthcoming in BIAS 2026 at ECML/PKDD.}}
\authorrunning{J.M. \'Alvarez}
\institute{Santander AI Lab, Madrid, Spain \\
\email{jm.alvarez.colmenares@gmail.com}
}

\maketitle

\begin{abstract}
Causal perception occurs when agents with competing Structural Causal Models (SCMs) of the same system infer different probability distributions, including the hypothetical distributions implied by each agent's SCM under the same set of interventions. It shapes how agents reason about the system and how they perceive its fairness. Causal perception is a promising probabilistic framework, but it has remained purely theoretical. This work provides the first implementation of the causal perception framework of \'{A}lvarez and Ruggieri~\cite{Alvarez2025CausalPerception}. We operationalize structural (agents disagree on the causal graph) and parametrical (agents agree on the causal graph but disagree on its weights) causal perception. We design algorithms for computing interventional and counterfactual distributions and propose suitable distance measures to quantify the disagreement. Using the German Credit dataset, we illustrate how causal perception affects accuracy and fairness in a multi-expert decision setting. We show that the perception verdict is sensitive to the choice of distance metric and threshold. We also show that causal perception changes fairness assessments and threshold-based decisions. Bias proves situated with respect to the agent's SCM, demonstrating that competing worldviews in fairness problems cannot be ignored.
\end{abstract}

% ============================================================
\section{Introduction}
\label{sec:Introduction}
% ============================================================

Responsible AI often relies on Structural Causal Models (SCMs) to reason about the fairness of a system \cite{DBLP:conf/nips/KusnerLRS17,DBLP:conf/aaai/Chiappa19,DBLP:conf/nips/JavaloySV23,DBLP:conf/fat/BaumannCCIR23,DBLP:journals/jair/AlvarezR25,DBLP:journals/corr/abs-2405-13693}. SCMs are graphical models that represent the causal relationships between variables as cause-effect pairs and allow us to reason about hypothetical ``what if'' scenarios in a system. Each SCM represents context-specific knowledge or assumptions, encoding individual or group beliefs on how the system works and, thus, summarizing a worldview. Multiple SCMs can represent the same system, but the literature typically relies on a single SCM. This practice is problematic because different SCMs can imply different fairness interpretations of the same system \cite{DBLP:conf/nips/RussellKLS17}. Dealing with multiple SCMs should be the norm rather than the exception, especially in settings in which competing worldviews are possible or even encouraged \cite{DBLP:conf/aaai/Ruggieri0PST23}. 

Recent work by \'Alvarez and Ruggieri \cite{Alvarez2025CausalPerception} tackles the problem of multiple and competing SCMs by introducing an SCM-based probabilistic framework to model perception. Perception refers to the cognitive phenomenon in which the same information is interpreted differently by individuals \cite{KahnemanNIPS2021}. \'Alvarez and Ruggieri \cite{Alvarez2025CausalPerception} repurpose the term for machine learning and introduce the notion of causal perception, or the difference in observed and/or hypothetical probability distribution(s) on the same information due to competing SCMs. Here, individuals interpret and reason about this information using probabilistic causal reasoning through individual-specific SCMs. Each SCM represents a worldview and conditions how each individual acts given a context. \'Alvarez and Ruggieri \cite{Alvarez2025CausalPerception} propose two kinds of causal perception: structural perception when individuals disagree on the cause-effect pairs (i.e., the graph) and parametrical perception when individuals agree on these pairs but disagree on the nature of the causal relationship (i.e., the weights). The framework is useful but remains theoretical, with its practical viability yet to be explored.

In this paper, we provide the first implementation of the causal perception framework. We operationalize both structural and parametrical causal perception, proposing a set of algorithms for the interventional and counterfactual levels of causal reasoning, with the observational level covered as the empty intervention. We illustrate the phenomenon on a real-world financial dataset using two model-based agents with competing SCMs, or worldviews, to examine how causal perception materializes and affects accurate and fair decision-making on the same pool of credit applicants (Example~\ref{ex:CreditScoring}). The experiments are illustrative, motivating the impact of causal perception in decision flows with multiple experts. We find, for instance, that two receivers differing on a single causal edge reach demographic parity gaps of $-0.594$ vs.\ $-0.170$ and disagree on 21.7\% of individual loan decisions. Our implementation is the first practical step toward applied causal perception research in AI systems.

\begin{example}
\label{ex:CreditScoring}
Consider two credit officers evaluating a pool of loan applicants. Officers have access to the applicants' gender, age, financial status, and credit information and use this information to make their decisions (see Fig.\@~\ref{fig:chiappa_dag}). Each officer decides the creditworthiness of an applicant, often relying on hypothetical scenarios to justify their decisions. 
% Officers are asked to monitor the fairness of their decisions. 
Suppose that one officer believes that gender affects the information considered, while the other officer believes that it does not. This divergence in reasoning materializes in competing SCMs and induces different hypothetical distributions about the same applicant pool and the perceived fairness of the decision-making process. This disagreement is precisely what the causal perception framework captures.
\end{example}

Our contributions are threefold. 
First, we operationalize causal perception for single- and multi-intervention settings at both the interventional and counterfactual levels of Pearl's causal ladder, providing two algorithms that translate the framework's definitions into executable procedures. 
Second, we illustrate both structural and parametrical causal perception on real data, revealing that the perception verdict is sensitive to the choice of distance metric and threshold. 
Third, we show that even minor SCM disagreements propagate into different fairness verdicts, substantiating the notion that bias is situated with respect to a receiver's worldview in the form of a SCM.

Moving forward, Section~\ref{sec:Background} introduces the background, Section~\ref{sec:Methodology} presents the causal perception implementation, Section~\ref{sec:Experiments} illustrates the implementation on real data, and Section~\ref{sec:Discussion} concludes this paper.

% ============================================================
\section{Background}
\label{sec:Background}
% ============================================================

\subsection{Preliminaries}

\paragraph{Structural Causal Models.} Let $P(\mathbf{X})$ denote the joint probability distribution of the set of random variables $\mathbf{X} = \{X_1, \ldots, X_p\}$. The SCM describes how these $p$ variables relate to each other as cause-effect pairs. Formally, an SCM $\mathcal{M}$ \cite{PearlCausality2009} is a tuple $\mathcal{M} = \langle \mathbf{U}, \mathbf{X}, \mathbf{F} \rangle$ that transforms a set of $p$ latent variables $\mathbf{U} \sim P(\mathbf{U})$ into a set of $p$ observed variables $\mathbf{X}$ based on a set of structural equations $\mathbf{F}$:
\begin{equation}
\label{eq:SCM}
    P(\mathbf{U}) = P(U_1, \dots, U_p) \;\;\; X_j := f_j(X_{pa(j)}, U_j), \quad j=1,\dots,p
\end{equation}
where $U_j \in \mathbf{U}$, $X_j \in \mathbf{X}$, and $f_j \in \mathbf{F}$. Each function $f_j$ maps the parents $X_{pa(j)}$ and the latent variable $U_j$ to $X_j$. Through $\mathcal{M}$ we can draw a causal graph $\mathcal{G}$ where each node represents a variable and each directed edge a causal relationship. For simplicity, we assume $\mathcal{G}$ to be a directed acyclic graph (DAG).

\paragraph{Probabilistic reasoning.} When reasoning about $P(\mathbf{X})$ given an SCM $\mathcal{M}$ (or a causal graph $\mathcal{G}$), we deal with observed (\textit{what is}) and hypothetical (\textit{what if} and \textit{what could have been}) distributions.\footnote{We assume that reasoning about $\mathbf{X}$ is probability-based, a common practice in fields that model human behavior like cognitive psychology \cite{tversky_judgment_1974} and machine learning \cite{PearlCausality2009}.} 
For the unique observed distribution, we use $\mathcal{M}$ to factor $P(\mathbf{X})$ into parent-child, or cause-effect, pairs:
\begin{equation}
\label{eq:MarkovCondition}
    P(\mathbf{X}) = \prod_{j=1}^{p} P(X_j \mid X_{pa(j)}).
\end{equation}
For the hypothetical distributions, we use $\mathcal{M}$ and the \textit{do-operator} $do(X_i=x_i)$, which replaces the structural equation in $\mathbf{F}$ for $X_i$ with the value $x_i$, to manipulate $P(\mathbf{X})$. Each intervention implies a new distribution $P(\mathbf{X})^{do(i)}$ for $i \in \mathcal{I}_{\mathbf{X}}$, called the $i$-interventional distribution, where $\mathcal{I}_{\mathbf{X}}$ denotes the set of all interventions. We represent these distributions as the poset of all interventional distributions implied by $\mathcal{M}$:
\begin{equation}
\label{eq:PosetofDist}
    \mathcal{P}_{\mathbf{X}} := \Big( \Big\{ P(\mathbf{X})^{do(i)}: i \in \mathcal{I}_{\mathbf{X}} \Big\}, \, \leq_{\mathbf{X}} \Big)
\end{equation}
where $\leq_{\mathbf{X}}$ is the natural partial ordering inherited from $\mathcal{I}_{\mathbf{X}}$ \cite{DBLP:conf/uai/RubensteinWBMJG17}. Note that $P(\mathbf{X}) \in \mathcal{P}_{\mathbf{X}}$ and that $\mathcal{P}_{\mathbf{X}}$ is a singleton comprising $P(\mathbf{X})$ when $\mathcal{I}_{\mathbf{X}} = \{ \emptyset \}$. Equation~\ref{eq:PosetofDist} represents all the possible ways we can reason about $P(\mathbf{X})$ given an SCM $\mathcal{M}$. It captures all three levels of Pearl's causal ladder: \textit{association} (1st rung), \textit{intervention} (2nd rung), and \textit{counterfactual} (3rd rung) \cite{Pearl2018WHy}.

\'Alvarez and Ruggieri \cite{Alvarez2025CausalPerception} discuss causal perception at two levels: observational and hypothetical. The latter summarizes the 2nd and 3rd rungs under hypothetical distributions. What separates these two levels is the additional abduction step performed in the 3rd rung in which we use the evidence, meaning the observed data, to infer the latent variables $\mathbf{U}$. Abduction implies approximating the posterior distribution $P(\mathbf{U} \mid \mathbf{X})$ and drawing $\hat{\mathbf{U}}$ at the individual level. Both 2nd and 3rd rungs rely on the same $i$-intervention (hence, the approach in \cite{Alvarez2025CausalPerception}). What changes is whether $\mathbf{U}$ is inferred and used to compute the distribution. 
In the 2nd rung, we intervene and propagate the downstream effects without inferring $\mathbf{U}$. In the 3rd rung, we do the same but adjust this propagation with individual-level $\hat{\mathbf{U}}$ drawn from $P(\mathbf{U} \mid \mathbf{X})$. 
Whether the manipulation is interventional or counterfactual is clear from the context. 

\paragraph{Causal perception.} Causal perception centers on the receiver $R \in \mathcal{R}$, where $\mathcal{R}$ is the set of all possible receivers, and the sender $S \in \mathcal{S}$, where $\mathcal{S}$ is the set of all possible senders. These are ``agents'', in the widest sense of the word, that receive and send, respectively, the information $\mathbf{X}$. Receivers, notably, interpret it as a probability distribution $P(\mathbf{X})$ with a corresponding SCM $\mathcal{M}$. 

Consider $R_i, R_j \in \mathcal{R}$ with SCMs $\mathcal{M}_{R_i}$ and $\mathcal{M}_{R_j}$ for information $\mathbf{X}$ provided by $S \in \mathcal{S}$.
% For a threshold $\epsilon \in \mathbb{R}^+$, 
Causal perception occurs when:
\begin{equation}
\label{eq:CausalPerception}
  \bar{d} \left( \mathcal{P}_{\mathbf{X}_{R_i}}, \mathcal{P}_{\mathbf{X}_{R_j}} \right) > \epsilon
\end{equation}
where $\mathcal{P}_{\mathbf{X}_{R_i}}$ and $\mathcal{P}_{\mathbf{X}_{R_j}}$ represent the poset of all distributions of $\mathbf{X}$ induced by $\mathcal{M}_{R_i}$ and $\mathcal{M}_{R_j}$; 
$\bar{d}(\cdot,\cdot)$ represents a suitable aggregated distance measure between two sets of probability distributions; and $\epsilon \in \mathbb{R}^+$ represents a threshold \cite{Alvarez2025CausalPerception}. 
% Similarly, 

We can also define causal perception on an \textit{l}-intervention $\mathcal{I}_{\mathbf{X}} = \{l\}$:
\begin{equation}
\label{eq:Perception2}
    d \left( P_{R_i}(\mathbf{X})^{do(l)}, P_{R_j}(\mathbf{X})^{do(l)} \right) > \epsilon 
\end{equation}
where we study the disagreement of the interventional distributions resulting from the \textit{l}-intervention via the SCMs $\mathcal{M}_{R_i}$ and $\mathcal{M}_{R_j}$. 

By treating $f_j$ as an additive noise model (ANM), two kinds of causal perception are proposed. 
% Both explain why the individual posets diverge by $\epsilon$. 
Formally, $f_j$ is a linear transformation such that $f_j := \sum_{i=1}^{|pa(j)|} \beta_{ij} \cdot X_{pa(j)_i} + U_j$ with $\beta_{ij}\in\mathbb{R}$ denoting the causal weight of the $i$-th parent $pa(j)_i$ of $X_j$. These are: \textbf{structural perception}, where agents disagree on the causal graph $\mathcal{G}$ for $P(\mathbf{X})$; and \textbf{parametrical perception}, where agents agree on the causal graph $\mathcal{G}$ for $P(\mathbf{X})$ but disagree on the causal weights $\beta_{ij}$.

\begin{remark}%[Worldviews]
% For our purposes, 
We use a broad interpretation of receivers $R_i$ and $R_j$: these can range from humans to machine learning models that, within a certain context, act based on their causal interpretation of $\mathbf{X}$. Further, we take for granted how the agents' SCMs are constructed and focus our efforts on the perception implementation aspect. Furthermore, we avoid the notion of a ground-truth SCM when it comes to human-based systems as these represent worldviews. 
\end{remark}

\paragraph{Situated bias.} \'Alvarez and Ruggieri \cite{Alvarez2025CausalPerception} also use causal perception to revisit the implicit stance that bias is objective, static, and universal. Most responsible AI methods speak of bias and, in turn, of fairness as a shared fact. In reality, however, this seems not to be the case as what is fair to some may not be fair to others \cite{DBLP:conf/kdd/SrivastavaHK19,DBLP:conf/aies/Yaghini0H21,DBLP:conf/aies/BertrandBEM22}. Bias is always situated as, at a minimum, it requires a reference worldview. Causal perception, thus, would apply to the context of bias once we rank one receiver's interpretation of $\mathbf{X}$ over the others. 
% Similarly, this stance allows for the possibility in which defining what bias is depends on multiple, competing worldviews.

% Under situated bias, w
We revisit three popular fairness metrics, demographic parity (DP) \cite{DBLP:conf/icdm/CaldersKP09}, equalized odds (EO) \cite{DBLP:conf/nips/HardtPNS16}, and counterfactual fairness (CF) \cite{DBLP:conf/nips/KusnerLRS17}, by situating each relative to a receiver (see Appendix~\ref{app:Supp_Material}). For instance, DP for $R_i$ would amount to $P_{R_i}(\hat{Y} = 1 \mid A = a') = P_{R_i}(\hat{Y} = 1 \mid A = a)$ with $A$ denoting the non-trivial binary protected attribute and $\hat{Y}$ denoting the predicted outcome by $R_i$. It introduces the possibility for a second $R_j$ to have its own reading of DP. In either case, we would say that the potential bias detected from DP would be situated relative to $R_i$ or $R_j$ as two receivers with different SCMs may satisfy or violate the same criterion. This would apply to other fairness metrics \cite{DBLP:conf/icse/VermaR18}. 

\subsection{Related Work}

% \paragraph{Additional related work.} 
Most causal fairness works focus on what might be missing in an SCM and how that impacts the CF \cite{DBLP:conf/nips/KusnerLRS17} assessment of a system. 
Works range from graph uncertainty \cite{DBLP:journals/corr/abs-2601-03203} and missing variables \cite{DBLP:conf/uai/KilbertusBKWS19} to misspecified functional forms \cite{DBLP:conf/nips/JavaloySV23}. 
The implicit focus is on, one, the existence of a ground-truth SCM to aim for and, two, consolidating all possible assessments on a single SCM. Russell et al.\ \cite{DBLP:conf/nips/RussellKLS17} is a good example of this mono-model approach centered on CF. Recent work \cite{DBLP:journals/corr/abs-2502-01211} goes beyond CF, but the focus remains on a single SCM as \textit{the} worldview.

Causal perception challenges this dominant approach.
% in causal fairness. 
It rejects the premise that competing SCMs must be consolidated, whether by selecting one or by fusing them into a consensus model, before any inference is drawn. 
Consolidation presupposes a single ground-truth SCM and treats disagreement as noise to average away. The concern echoes predictive multiplicity, where distinct models fit the same data equally well yet disagree on individual predictions \cite{DBLP:journals/statsci/Breiman01,DBLP:conf/icml/MarxCU20}.

% Causal perception 
The framework instead addresses the complementary setting in which the disagreement itself is the object of study. In multi-expert decision flows, like Example~\ref{ex:CreditScoring}, each expert acts on their own worldview rather than on a consensus model. Fusion asks which model, or combination of models, is best; perception quantifies what is at stake when the models differ. Previous work \cite{DBLP:conf/kdd/SrivastavaHK19,DBLP:conf/aies/Yaghini0H21,DBLP:conf/aies/BertrandBEM22} raises concerns around assuming single, shared notions of fairness in practice. \'Alvarez and Ruggieri \cite{Alvarez2025CausalPerception} extend this view to the causal realm, embracing the setting of competing worldviews. We extend this line of work by implementing causal perception for the first time.

% ============================================================
\section{Implementing Causal Perception}
\label{sec:Methodology}
% ============================================================

% We detail the implementation of causal perception. 
Causal perception requires a set of interventions, a distance measure, and a threshold.
Given the individual SCMs, it reduces to measuring distributional divergence under a chosen set of interventions. 
Importantly, receivers share the variable set $\mathbf{X}$ and, thus, any disagreement in causal perception is about the edges, not the nodes.

\subsection{Algorithms}

\begin{algorithm}[t]
\caption{Single-Intervention Causal Perception}\label{alg:single_intervention}
\begin{algorithmic}[1]
\Require SCMs $\mathcal{M}_{R_i} = \langle \mathbf{U}_i, \mathbf{X}, \mathbf{F}_i \rangle$ and $\mathcal{M}_{R_j} = \langle \mathbf{U}_j, \mathbf{X}, \mathbf{F}_j \rangle$
\Require Intervention $l \in \mathcal{I}_{\mathbf{X}}$ such that $l = do(X_k = x_k)$
\Require Distance function $d(\cdot, \cdot)$, threshold $\epsilon \in \mathbb{R}^+$
\Require Rung $r \in \{2,3\}$; if $r{=}3$: factual evidence $\mathbf{x}_{\text{obs}}$
\Ensure Distance $\delta \in \mathbb{R}^+$, perception flag $\phi \in \{0,1\}$
\If{$r = 3$} \Comment{Abduction (3rd rung only)}
    \State Compute $P(\mathbf{U}_i \mid \mathbf{X}{=}\mathbf{x}_{\text{obs}})$ from $\mathcal{M}_{R_i}$ and $P(\mathbf{U}_j \mid \mathbf{X}{=}\mathbf{x}_{\text{obs}})$ from $\mathcal{M}_{R_j}$
\EndIf
\State Replace $f_k \in \mathbf{F}_i$ and $f_k \in \mathbf{F}_j$ with $X_k := x_k$ \Comment{Action}
\If{$r = 2$} \Comment{Prediction}
    \State Compute $P_{R_i}(\mathbf{X})^{do(l)}$ and $P_{R_j}(\mathbf{X})^{do(l)}$
\Else
    \State Compute $P_{R_i}(\mathbf{X}(\hat{\mathbf{U}}_i))^{do(l)}$ and $P_{R_j}(\mathbf{X}(\hat{\mathbf{U}}_j))^{do(l)}$
\EndIf
\State $\delta \gets d\!\left( P_{R_i}^{do(l)},\; P_{R_j}^{do(l)} \right)$
\State $\phi \gets \mathbf{1}[\delta > \epsilon]$
\State \Return $(\delta, \phi)$
\end{algorithmic}
\end{algorithm}
\begin{algorithm}[t]
\caption{Multi-Intervention Causal Perception}\label{alg:multi_intervention}
\begin{algorithmic}[1]
\Require SCMs $\mathcal{M}_{R_i} = \langle \mathbf{U}_i, \mathbf{X}, \mathbf{F}_i \rangle$ and $\mathcal{M}_{R_j} = \langle \mathbf{U}_j, \mathbf{X}, \mathbf{F}_j \rangle$
\Require Set of interventions $\mathcal{I}_{\mathbf{X}}$
\Require Distance function $d(\cdot, \cdot)$, aggregation function $\bar{d}$, threshold $\epsilon \in \mathbb{R}^+$
\Require Rung $r \in \{2,3\}$; if $r{=}3$: factual evidence $\mathbf{x}_{\text{obs}}$
\Ensure Aggregated distance $\bar{\delta} \in \mathbb{R}^+$, per-intervention distances $\boldsymbol{\Delta}$, perception flag $\phi \in \{0,1\}$
\State $\boldsymbol{\Delta} \gets \{\}$
\For{each $l \in \mathcal{I}_{\mathbf{X}}$}
    \State $(\delta_l, \_) \gets$ \Call{Algorithm~\ref{alg:single_intervention}}{$\mathcal{M}_{R_i}, \mathcal{M}_{R_j}, l, d, \epsilon, r$}
    \State $\boldsymbol{\Delta} \gets \boldsymbol{\Delta} \cup \{\delta_l\}$
\EndFor
\State $\bar{\delta} \gets \bar{d}(\boldsymbol{\Delta})$
\State $\phi \gets \mathbf{1}[\bar{\delta} > \epsilon]$
\State \Return $(\bar{\delta}, \boldsymbol{\Delta}, \phi)$
\end{algorithmic}
\end{algorithm}

Algorithm~\ref{alg:single_intervention} operationalizes causal perception for a single $l$-intervention (Equation~\ref{eq:Perception2}). It handles both the interventional (2nd rung) and counterfactual (3rd rung) levels of Pearl's causal ladder \cite{Pearl2018WHy} via the rung parameter $r$. When $r{=}3$, an abduction step infers the latent $\mathbf{U}$ from the observed data before intervening. Algorithm~\ref{alg:multi_intervention} extends this to the multi-intervention case (Equation~\ref{eq:CausalPerception}), aggregating per-intervention distances into a single measure via $\bar{d}$. Both algorithms are direct translations of Equations~\ref{eq:Perception2} and~\ref{eq:CausalPerception}. 

The implementation of Algorithm~\ref{alg:single_intervention} is straightforward. It depends on a single manipulation over the SCMs. Algorithm~\ref{alg:multi_intervention}, instead, can become computationally expensive. 
For instance, if all variables in $\mathbf{X}$ are binary, then there are at least $2^p$ possible interventions. This can become infeasible for large $p$ or when variables are continuous. In practice, rather than considering the full set of interventions $\mathcal{I}_{\mathbf{X}}$, we consider a subset of interventions that are most relevant for the analysis of interest, $\mathcal{I}_{\mathbf{X}}^* \subset \mathcal{I}_{\mathbf{X}}$, based on domain knowledge or prior assumptions. We leave the development of more efficient algorithms for future work.

The nature of $\mathcal{I}_{\mathbf{X}}^*$ will depend on which variables to manipulate (variable selection) and how to manipulate them (value selection). For the former, we may want to intervene on a subset of variables (for example, \texttt{gender} and \texttt{income}). For the latter, we may want to intervene on specific values of the variables (\texttt{gender = male} and \texttt{income = avg. income}). Following the ordering induced by $\mathcal{I}_{\mathbf{X}}$ \cite{DBLP:conf/uai/RubensteinWBMJG17}, interventions inherit the structure of the SCM.

\subsection{Distance Measure}

The framework places no restriction on $d(\cdot,\cdot)$. We consider three distance measures commonly used in machine learning \cite{DBLP:journals/kais/GoldenbergW19}. Each distance captures a different notion of distributional divergence.

The \textit{Wasserstein-2 distance} $W_2(P, Q)$ \cite{Villani2009OptimalTransport} is defined for multivariate distributions $P$ and $Q$ over $\mathbb{R}^p$ as:
\begin{equation}
\label{eq:Wasserstein2}
    W_2(P, Q) = \left( \inf_{\gamma \in \Gamma(P,Q)} \int \| \mathbf{x} - \mathbf{y} \|^2 \, d\gamma(\mathbf{x}, \mathbf{y}) \right)^{1/2}
\end{equation}
where $\Gamma(P,Q)$ denotes the set of all couplings of $P$ and $Q$. It captures both location and shape differences between distributions. It measures the minimum ``cost'' of transporting probability mass from one distribution to the other.

The \textit{Kullback-Leibler divergence} $D_{\mathrm{KL}}(P \| Q)$ \cite{KullbackLeibler1951} measures the information lost when $Q$ is used to approximate $P$:
\begin{equation}
\label{eq:KL}
    D_{\mathrm{KL}}(P \| Q) = \int p(\mathbf{x}) \log \frac{p(\mathbf{x})}{q(\mathbf{x})} \, d\mathbf{x}.
\end{equation}
Unlike $W_2$, $D_{\mathrm{KL}}$ is asymmetric and unbounded; it is the standard information-theoretic baseline. It quantifies how ``surprised'' an observer would be if they expected data from $Q$ but instead observed data from $P$. We include it as a robustness check capturing a different notion of divergence: distributional surprise rather than geometric displacement.

The \textit{total variation distance} $\mathrm{TV}(P, Q)$ \cite{Tsybakov2009NonparametricEstimation} provides the maximum difference in probability assigned to any event based on $P$ and $Q$:
\begin{equation}
\label{eq:TV}
    \mathrm{TV}(P, Q) = \frac{1}{2} \int | p(\mathbf{x}) - q(\mathbf{x}) | \, d\mathbf{x}.
\end{equation}
$\mathrm{TV}$ is bounded in $[0,1]$, giving a normalized, probability-scale measure. It answers the question: for the event on which $P$ and $Q$ disagree the most, how large is that disagreement? This is useful for interpreting the threshold $\epsilon$: a value of $\mathrm{TV} = 0.1$ means the two distributions differ by at most $10\%$ on any event.

\subsection{Aggregation Function and Threshold Choice}

For the multi-intervention case (Algorithm~\ref{alg:multi_intervention}), the aggregation function $\bar{d}$ summarizes the per-intervention distances $\boldsymbol{\Delta} = \{ \delta_l : l \in \mathcal{I}_{\mathbf{X}} \}$ into a scalar. Natural candidates include the mean $\bar{d}(\boldsymbol{\Delta}) = \frac{1}{|\mathcal{I}_{\mathbf{X}}|} \sum_{l} \delta_l$, the maximum $\bar{d}(\boldsymbol{\Delta}) = \max_{l} \delta_l$, or a weighted average that assigns more importance to certain interventions. 

The choice of $\bar{d}$ is normative. For instance, the mean captures average-case disagreement, the maximum captures worst-case disagreement, and a weighted average reflects domain priorities. We adopt the mean as our default and, abusing notation slightly, write $\bar{d}(W_2)$ for the aggregated $W_2$ distance (similarly for $D_{\mathrm{KL}}$ and $\mathrm{TV}$). The aggregate complements, rather than replaces, per-intervention reasoning: Algorithm~\ref{alg:multi_intervention} returns the per-intervention distances $\boldsymbol{\Delta}$ alongside $\bar{\delta}$. The aggregate serves questions that need a single system-level verdict, such as auditing whether two receivers perceive the same system differently, while $\boldsymbol{\Delta}$ supports reasoning on specific interventions.

The threshold $\epsilon$ determines when two receivers are deemed to perceive differently (Equations~\ref{eq:CausalPerception} and~\ref{eq:Perception2}). Rather than fixing $\epsilon$ a priori, we treat it as a parameter to be swept: for a range of $\epsilon$ values, we classify each SCM pair as exhibiting causal perception ($\phi = 1$) or not ($\phi = 0$). This sweep reveals the sensitivity of the classification to the tolerance level.

% ============================================================
\section{Illustration: German Credit}
\label{sec:Experiments}
% ============================================================

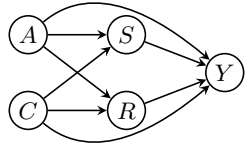
\begin{figure}[t]
\centering
\begin{tikzpicture}[
    >=stealth,
    every node/.style={draw, circle, minimum size=0.5cm, inner sep=0.5pt, font=\footnotesize},
    semithick
]
  \node (A) at (0, 0.5) {$A$};
  \node (C) at (0,-0.5) {$C$};
  \node (S) at (1.3, 0.5) {$S$};
  \node (R) at (1.3,-0.5) {$R$};
  \node (Y) at (2.6, 0) {$Y$};
  \draw[->] (A) -- (S);
  \draw[->] (A) -- (R);
  \draw[->] (A) to[out=40,in=140] (Y);
  \draw[->] (C) -- (S);
  \draw[->] (C) -- (R);
  \draw[->] (C) to[out=-40,in=220] (Y);
  \draw[->] (S) -- (Y);
  \draw[->] (R) -- (Y);
\end{tikzpicture}
\caption{
% Causal graph for German Credit from Chiappa \cite{DBLP:conf/aaai/Chiappa19}. 
$A$: gender; $C$: age; $S$: status of checking account, savings, and housing; $R$: credit amount and repayment duration; $Y$: good or bad credit risk \cite{DBLP:conf/aaai/Chiappa19}.}
\label{fig:chiappa_dag}
\end{figure}

We illustrate causal perception using the German Credit dataset.\footnote{\url{https://github.com/SantanderAI/causal-perception-implementation}.} It contains 20 attributes of 1,000 loan applicants: each applicant is classified as a good ($Y=1$) or bad ($Y=0$) credit risk. 
We adopt the causal graph in Fig.\@~\ref{fig:chiappa_dag} from Chiappa~\cite{DBLP:conf/aaai/Chiappa19}, focusing on eight of these attributes: $A$ represents gender, $C$ age, $S$ status of checking account, savings, and housing, $R$ credit amount and repayment duration, and $Y$ credit classification. Gender is derived from the dataset's personal-status attribute, which conflates sex and marital status; we follow Chiappa's preprocessing convention \cite{DBLP:conf/aaai/Chiappa19}.

We use a 70\%--30\% train-test split. 
We set $\epsilon = 0.1$ as an illustrative threshold. Since the three metrics live on different scales, $\epsilon$ is not a universal calibration; each distance may require a metric-specific threshold.
Since $\phi = 1$ exactly when a distance exceeds $\epsilon$, each reported distance also summarizes the full threshold sweep: perception is flagged for every $\epsilon$ below the reported value and not flagged above it.  
We estimate $W_2$ with the exact one-dimensional quantile formula on paired samples, $D_{\mathrm{KL}}$ via Gaussian kernel density estimation with Scott's bandwidth, and $\mathrm{TV}$ via histograms with 50 common-support bins; the latter two estimates are sensitive to bandwidth and binning choices. 

Distances are accompanied by 95\% bootstrap confidence intervals ($B{=}1000$, percentile method, paired resampling of test-set indices). Because these are percentile bootstrap intervals, they need not contain the full-sample point estimate. In each experiment, we instantiate two model-based receivers $R_1$ and $R_2$, each holding a competing SCM, who evaluate the same 300 test applicants. 
We stress that these receivers are idealized computational agents rather than observed human decision-makers. We focus on the binary protected attribute $A$ ($0$ for female, $1$ for male) when looking at $\mathcal{I}^*_{\mathbf{X}}$. 
We focus on the predicted probability of good credit risk and the predicted risk classification.
We provide additional experiments in Appendix~\ref{app:Add_Material}.

\subsection{Structural Causal Perception}
\label{sec:structural_perception}

\begin{table}[t]
\centering
\caption{Distances for structural perception around $A$. Top panel: interventional (2nd rung) distributions. Bottom panel: counterfactual (3rd rung) distributions. Bracketed values are 95\% bootstrap confidence intervals.}
\label{tab:structural_perception}
\resizebox{\textwidth}{!}{%
\begin{tabular}{llccc}
\toprule
Comparison & Intervention & $W_2$ & $D_{\mathrm{KL}}$ & TV \\
\midrule
\multicolumn{5}{l}{\textit{Interventional (2nd rung)}} \\
$\mathcal{M}_1$ vs.\ $\mathcal{M}_2$ & $\mathrm{do}(A{=}0)$ & 0.052\;{\scriptsize[.052,.052]} & 0.911\;{\scriptsize[.774,1.12]} & 0.460\;{\scriptsize[.443,.550]} \\
$\mathcal{M}_1$ vs.\ $\mathcal{M}_2$ & $\mathrm{do}(A{=}1)$ & 0.026\;{\scriptsize[.025,.027]} & 0.272\;{\scriptsize[.230,.332]} & 0.397\;{\scriptsize[.387,.520]} \\
$\mathcal{M}_1$ vs.\ $\mathcal{M}_2$ & $\bar{d}$ (mean) & 0.039\;{\scriptsize[.038,.040]} & 0.591\;{\scriptsize[.502,.727]} & 0.428\;{\scriptsize[.415,.535]} \\
\midrule
\multicolumn{5}{l}{\textit{Counterfactual (3rd rung)}} \\
$\mathcal{M}_1$ vs.\ $\mathcal{M}_2$ & $\mathrm{do}(A{=}0)$ & 0.050\;{\scriptsize[.049,.052]} & 0.063\;{\scriptsize[.058,.087]} & 0.243\;{\scriptsize[.260,.387]} \\
$\mathcal{M}_1$ vs.\ $\mathcal{M}_2$ & $\mathrm{do}(A{=}1)$ & 0.024\;{\scriptsize[.023,.026]} & 0.014\;{\scriptsize[.012,.027]} & 0.200\;{\scriptsize[.217,.340]} \\
$\mathcal{M}_1$ vs.\ $\mathcal{M}_2$ & $\bar{d}$ (mean) & 0.037\;{\scriptsize[.036,.039]} & 0.039\;{\scriptsize[.035,.057]} & 0.222\;{\scriptsize[.238,.363]} \\
\bottomrule
\end{tabular}}%
\end{table}
\begin{figure}[t]
\centering
\includegraphics[width=\textwidth]{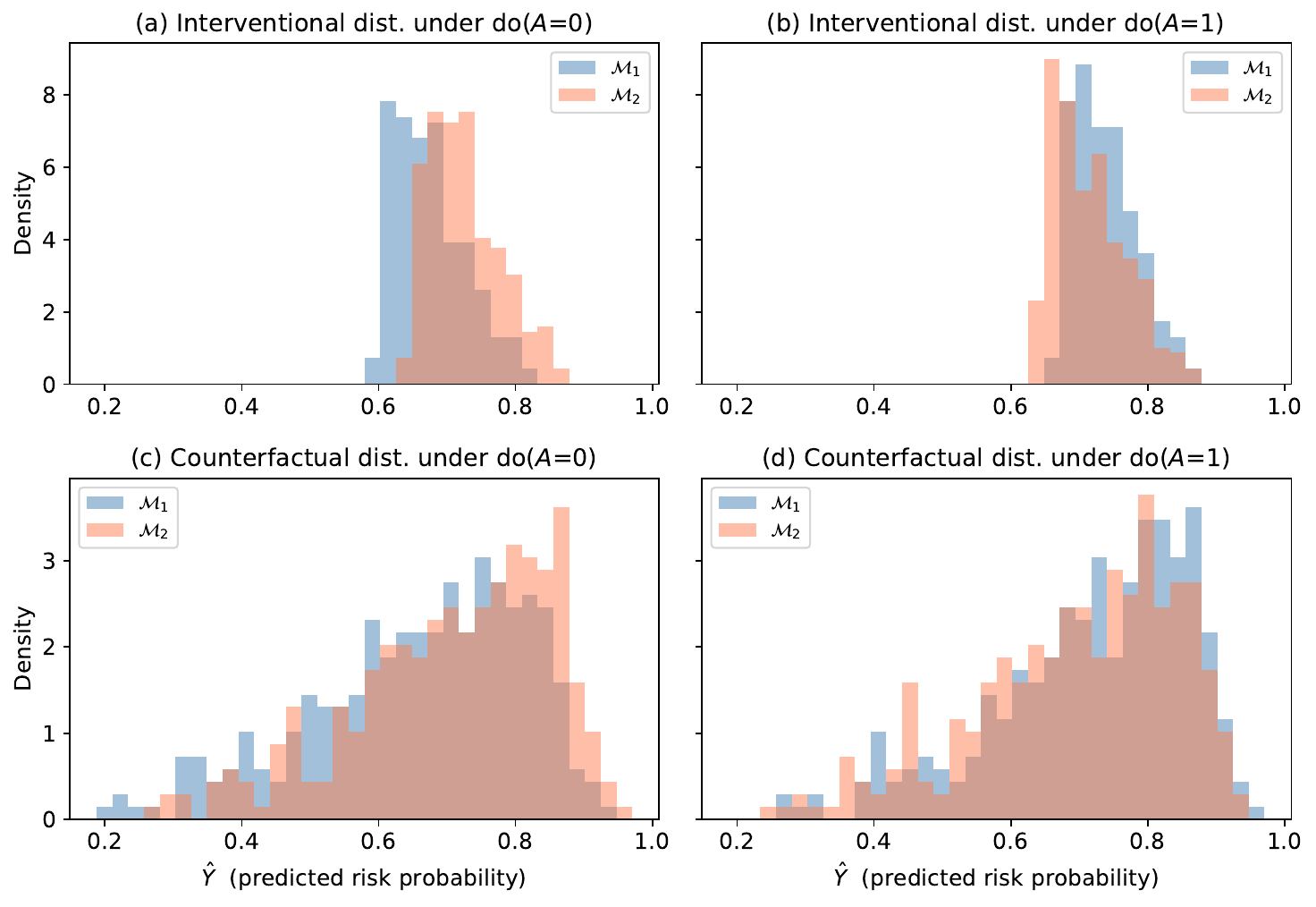}
\caption{Hypothetical distributions of predicted risk probability for structural causal perception. Top row: interventional (2nd rung). Bottom row: counterfactual (3rd rung).}
\label{fig:structural_perception}
\end{figure}

Under structural perception, the two receivers disagree on the causal graph. 
Suppose $R_1$ specifies the graph in Fig.\@~\ref{fig:chiappa_dag}, while $R_2$ specifies the same graph except that the edge $A \to Y$ is removed. In other words, $R_1$ and $R_2$ disagree on the direct effect of gender on credit risk. For each causal graph $\mathcal{G}_1$ and $\mathcal{G}_2$, we estimate the structural equations for each SCM $\mathcal{M}_1$ and $\mathcal{M}_2$ by fitting ordinary least squares models for continuous variables and logistic regression models for binary variables on the training data. For the counterfactual computations, abduction is residual-based, which is exact for linear ANMs: the mediator noise terms are recovered as regression residuals.  
% The outcome noise $U_Y$ is not abducted; the counterfactual credit risk is the model probability given counterfactual parent values.  

Table~\ref{tab:structural_perception} reports the distances for interventional (top panel) and counterfactual (bottom panel) distributions of the predicted credit risk, focusing on intervening on the protected attribute $A$. For both rungs, we compare the single-intervention setting for $A=1$ and $A=0$ (Algorithm~\ref{alg:single_intervention}) as well as the multi-intervention setting for $A$ as a whole (Algorithm~\ref{alg:multi_intervention}). Fig.\@~\ref{fig:structural_perception} visualizes the distributions. These represent settings in which the loan officers arrive at different ``what if'' and ``what could have been'' hypothetical distributions of credit risk about the applicant pool being male or female. 
% They both have access to the same data and corresponding set of attributes.  

The results reveal moderate divergence between the two SCMs at both rungs. At the \textit{interventional level}, the aggregated distance is $\bar{d}(W_2) = 0.039$, with the female intervention ($\mathrm{do}(A{=}0)$, $W_2 = 0.052$) producing a larger gap than the male one ($\mathrm{do}(A{=}1)$, $W_2 = 0.026$). Under $\epsilon = 0.1$, the $W_2$ distance does not detect structural causal perception, whereas $D_{\mathrm{KL}}$ and $TV$ do. This disagreement across metrics arises because the interventional distributions are narrow and slightly shifted in location: even a modest shift produces poor tail overlap and non-coinciding peaks, inflating $D_{\mathrm{KL}}$ and $TV$ relative to $W_2$. The perception verdict therefore depends on which aspect of distributional divergence the chosen metric emphasizes. Reading the sweep off Table~\ref{tab:structural_perception}: at the interventional level, $W_2$ flags perception only for $\epsilon < 0.039$, whereas $D_{\mathrm{KL}}$ and $\mathrm{TV}$ flag it for any $\epsilon$ up to $0.591$ and $0.428$, respectively.
At the \textit{counterfactual level}, the pattern is preserved but attenuated: $\bar{d}(W_2) = 0.037$, and only $TV$ exceeds $\epsilon$. The attenuation reflects the fact that the abduction step preserves individual heterogeneity in the mediators $S$ and $R$ via the inferred noise terms, producing wider distributions that absorb the between-model shift. Despite this, the rank ordering of per-intervention distances is unchanged, confirming that the structural disagreement on $A \to Y$ manifests consistently across both rungs of the causal ladder. 
We repeat the structural exercise for disagreement on $C \to Y$ in Appendix~\ref{app:structural_age}. The age experiment shows that causal perception is not specific to direct disagreement about gender. We also show that the pattern is robust to using a nonlinear outcome model in Appendix~\ref{app:nonlinear}, using a GBM instead of a logistic regression.

\subsection{Parametrical Causal Perception}

\begin{table}[t]
\centering
\caption{Distances for parametrical perception under $\mathrm{do}(A{=}1)$ (male). Distances under $\mathrm{do}(A{=}0)$ are zero and omitted. Top panel: interventional. Bottom panel: counterfactual. Bracketed values are 95\% bootstrap confidence intervals.}
\label{tab:parametrical_perception}
\resizebox{\textwidth}{!}{%
\begin{tabular}{llccc}
\toprule
Comparison & Intervention & $W_2$ & $D_{\mathrm{KL}}$ & TV \\
\midrule
\multicolumn{5}{l}{\textit{Interventional (2nd rung)}} \\
$\mathcal{M}_1$ vs.\ $\mathcal{M}_2$ & $\mathrm{do}(A{=}1)$ & 0.079\;{\scriptsize[.078,.080]} & 1.098\;{\scriptsize[1.002,1.29]} & 0.653\;{\scriptsize[.610,.717]} \\
\midrule
\multicolumn{5}{l}{\textit{Counterfactual (3rd rung)}} \\
$\mathcal{M}_1$ vs.\ $\mathcal{M}_2$ & $\mathrm{do}(A{=}1)$ & 0.076\;{\scriptsize[.074,.077]} & 0.115\;{\scriptsize[.106,.155]} & 0.257\;{\scriptsize[.277,.390]} \\
\bottomrule
\end{tabular}}%
\end{table}

Under parametrical perception, the two receivers agree on the causal graph but disagree on its causal weights. Suppose both $R_1$ and $R_2$ adopt the causal graph in Fig.\@~\ref{fig:chiappa_dag}. Here, to obtain two competing SCMs, we first estimate an SCM under this causal graph using the training set. Receiver $R_1$ holds this baseline SCM, $\mathcal{M}_1$. For $R_2$'s SCM $\mathcal{M}_2$, we again focus on the edges from $A$ to other variables. In line with the previous section, we consider the edge $A \rightarrow Y$. There are multiple ways to define the second SCM $\mathcal{M}_2$. We perturb the estimated causal weight of $A \rightarrow Y$ in $\mathcal{M}_1$ ($\hat{\beta}=0.383$) by drawing plausible values from its confidence interval.
For SCM $\mathcal{M}_2$, we consider $\hat{\beta} - 2\mathrm{SE}$, where $\mathrm{SE}$ is the standard error of the estimated causal weight, leading to a causal weight of 0.003.

Table~\ref{tab:parametrical_perception} reports the distances between the distributions at the interventional and counterfactual levels. 
Because the perturbed coefficient multiplies $A$, setting $A{=}0$ nullifies the perturbation: all distances under $\mathrm{do}(A{=}0)$ are exactly zero and are omitted. 
The divergence therefore concentrates entirely on $\mathrm{do}(A{=}1)$. We do not plot the distributions.
At the \textit{interventional level}, $W_2 = 0.079$, which does not exceed $\epsilon = 0.1$; $D_{\mathrm{KL}}$ and $TV$, however, both exceed the threshold, echoing the metric-sensitivity observed under structural perception. At the \textit{counterfactual level}, $W_2 = 0.076$ remains comparable, but both $D_{\mathrm{KL}}$ and $TV$ drop substantially ($D_{\mathrm{KL}}$: $1.098 \to 0.115$; $TV$: $0.653 \to 0.257$), and only $TV$ still exceeds $\epsilon$. As in the structural case, the abduction step widens the distributions and absorbs part of the between-model shift, attenuating all divergence measures. The overall pattern confirms that even a subtle coefficient change (here, within the 95\% confidence interval of the same estimator) can trigger parametrical causal perception under certain metrics, and that the counterfactual rung consistently dampens the signal relative to its interventional counterpart.

\subsection{On Accurate and Fair Decisions}

% Given the experimental results thus far, in which we have presented minor structural and parametrical changes between receivers $R_1$ and $R_2$, we now illustrate the implications for accurate and fair decision-making. 
We focus again on the test set, where we now use the ground-truth labels $Y$. For context, we consider the two credit officers (i.e., our $R_1$ and $R_2$) from Example~\ref{ex:CreditScoring}.
We consider structural causal perception as a way to understand how different SCMs lead to different decisions even when applied to the same data. Receiver $R_1$ holds $\mathcal{M}_1$ (Fig.\@~\ref{fig:chiappa_dag}); $R_2$ holds $\mathcal{M}_2$, which excludes the edge $A \to Y$. Both models are estimated on the training set. We are not interested in whether the internal models of these loan officers are the fairest or most accurate standalone models, but that they disagree with each other and can lead to different accuracy and fairness assessments.

\begin{figure}[t]
\centering
\includegraphics[width=\textwidth]{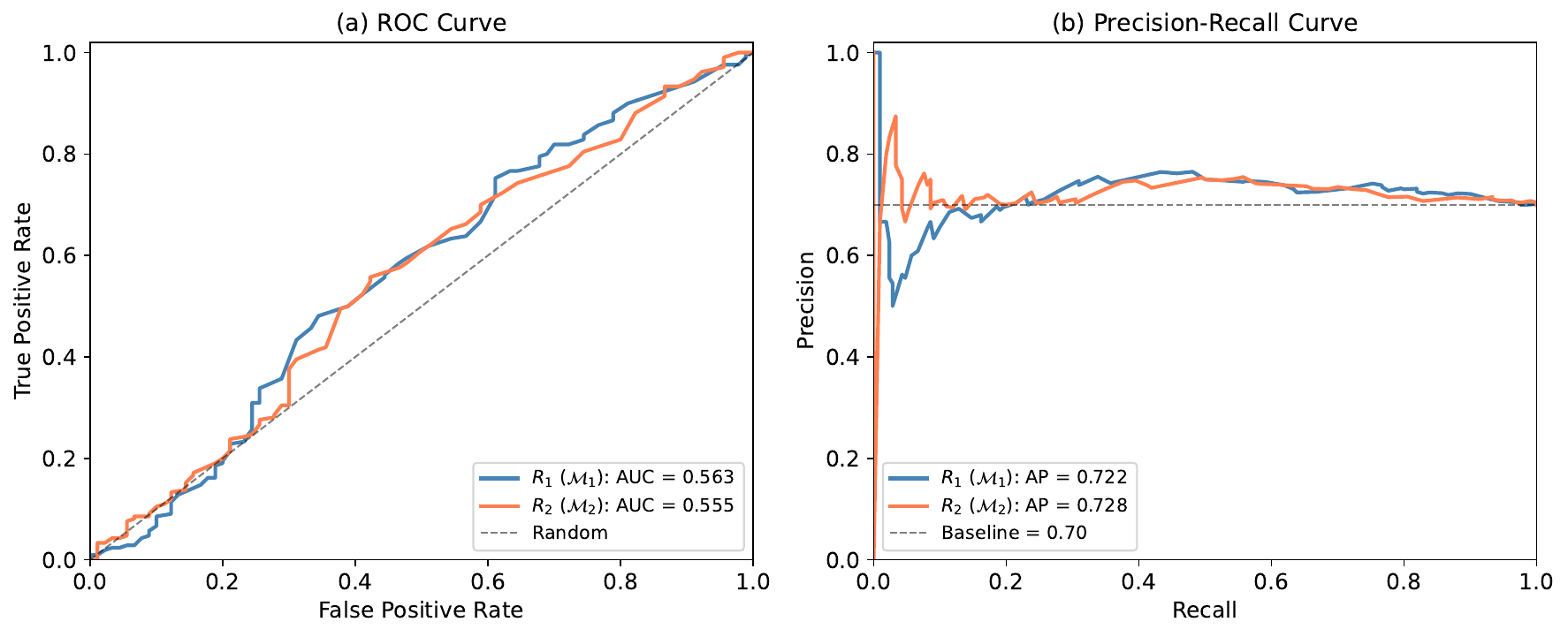}
\caption{ROC and PR curves for factual predictions of two receivers under structural causal perception: $\mathcal{G}_1$ as in Fig.\@~\ref{fig:chiappa_dag} and $\mathcal{G}_2$ without the edge $A \to Y$.}
\label{fig:accuracy}
\end{figure}

\paragraph{Accuracy.} We use each receiver's SCM to produce factual predicted risk probabilities $\hat{Y}$ and compare them against the ground-truth labels $Y$.\footnote{Factual predictions are obtained by propagating the test data through the SCM. We speak of ``factual'' because no manipulations are performed on the SCMs.} Since the test set is imbalanced (70\% good credit), we report both the ROC AUC and the Precision-Recall AUC (average precision). Fig.\@~\ref{fig:accuracy} shows the corresponding curves as we discretize the predicted risk probabilities over a range of thresholds. 

Receivers achieve near-identical ranking performance: ROC AUC is 0.563 for $R_1$ and 0.555 for $R_2$. The PR AUC values, 0.722 and 0.728, appear higher but should be read against the 0.70 base rate. The gap confirms that the presence or absence of the direct edge $A \to Y$ has a negligible impact on predictive accuracy. This modest performance is expected: the models are linear ANMs, not optimized classifiers. Structural perception, in this setting, does not affect predictive ranking performance.

\paragraph{Fairness.} Do the two receivers also reach similar fairness assessments? Let us first focus again on the \textit{factual predicted distributions} and apply a decision rule: grant the loan if $\hat{Y} \geq \tau$. Given the imbalanced test set (70\% good credit), we set $\tau = 0.70$ so that the threshold coincides with the base rate. Table~\ref{tab:fairness} reports DP~\cite{DBLP:conf/icdm/CaldersKP09} and EO~\cite{DBLP:conf/nips/HardtPNS16} metrics for each receiver around $A$, as well as the decision disagreement rate between $R_1$ and $R_2$.

\begin{table}[t]
\centering
\caption{Fairness metrics for the receivers based on factual predictions with $\tau = 0.70$. The DP gap is the difference in acceptance rates between female and male applicants; TPR and FPR gaps are the corresponding differences in true positive and false positive rates. Negative values indicate that females are disadvantaged.}
\label{tab:fairness}
\begin{tabular}{lccccc}
\toprule
 & \multicolumn{2}{c}{Accept.\ rate} & & \multicolumn{2}{c}{EO gaps} \\
\cmidrule{2-3} \cmidrule{5-6}
Receiver & Female & Male & DP gap & TPR gap & FPR gap \\
\midrule
$R_1$ ($\mathcal{M}_1$) & 0.240 & 0.833 & $-0.594$ & $-0.567$ & $-0.643$ \\
$R_2$ ($\mathcal{M}_2$) & 0.438 & 0.608 & $-0.170$ & $-0.134$ & $-0.228$ \\
\bottomrule
\end{tabular}
\end{table}
\begin{table}[t]
\centering
\caption{Within-model distances for counterfactual distributions (i.e., a population-level counterfactual sensitivity summary across applicants) for the receivers. Bracketed values are 95\% bootstrap confidence intervals.}
\label{tab:within_model}
\resizebox{\textwidth}{!}{%
\begin{tabular}{llccc}
\toprule
Receiver & 3rd Rung & $W_2$ & $D_{\mathrm{KL}}$ & TV \\
\midrule
$R_1$ ($\mathcal{M}_1$) & $\mathrm{do}(A{=}0)$ vs.\ $\mathrm{do}(A{=}1)$ & 0.061\;{\scriptsize[.060,.063]} & 0.088\;{\scriptsize[.081,.123]} & 0.207\;{\scriptsize[.247,.360]} \\
$R_2$ ($\mathcal{M}_2$) & $\mathrm{do}(A{=}0)$ vs.\ $\mathrm{do}(A{=}1)$ & 0.013\;{\scriptsize[.012,.013]} & 0.004\;{\scriptsize[.003,.008]} & 0.173\;{\scriptsize[.197,.320]} \\
\bottomrule
\end{tabular}}%
\end{table}

The two receivers reach different fairness verdicts. Under $R_1$, which includes the direct edge $A \to Y$, only 24\% of female applicants are approved compared to 83.3\% of males, yielding a DP gap of $-0.594$. The EO gaps are equally large: the true positive rate (TPR) gap ($-0.567$) and the false positive rate (FPR) gap ($-0.643$) indicate that $R_1$'s decisions systematically disadvantage women on both correctly and incorrectly classified applicants. Under $R_2$, which removes $A \to Y$, the gender disparity is less pronounced: acceptance rates are 43.8\% (female) vs.\ 60.8\% (male), with a DP gap of $-0.170$. The EO gaps follow the same pattern as DP.
Notably, 65 of the 300 applicants (21.7\%) receive a different loan decision depending on which receiver processes their application.
Together, these results show that causal perception is not merely a distributional abstraction: it translates into different decisions for the same individuals at the factual level. It also changes fairness assessments of the same system.

\begin{figure}[t]
\centering
\includegraphics[width=\textwidth]{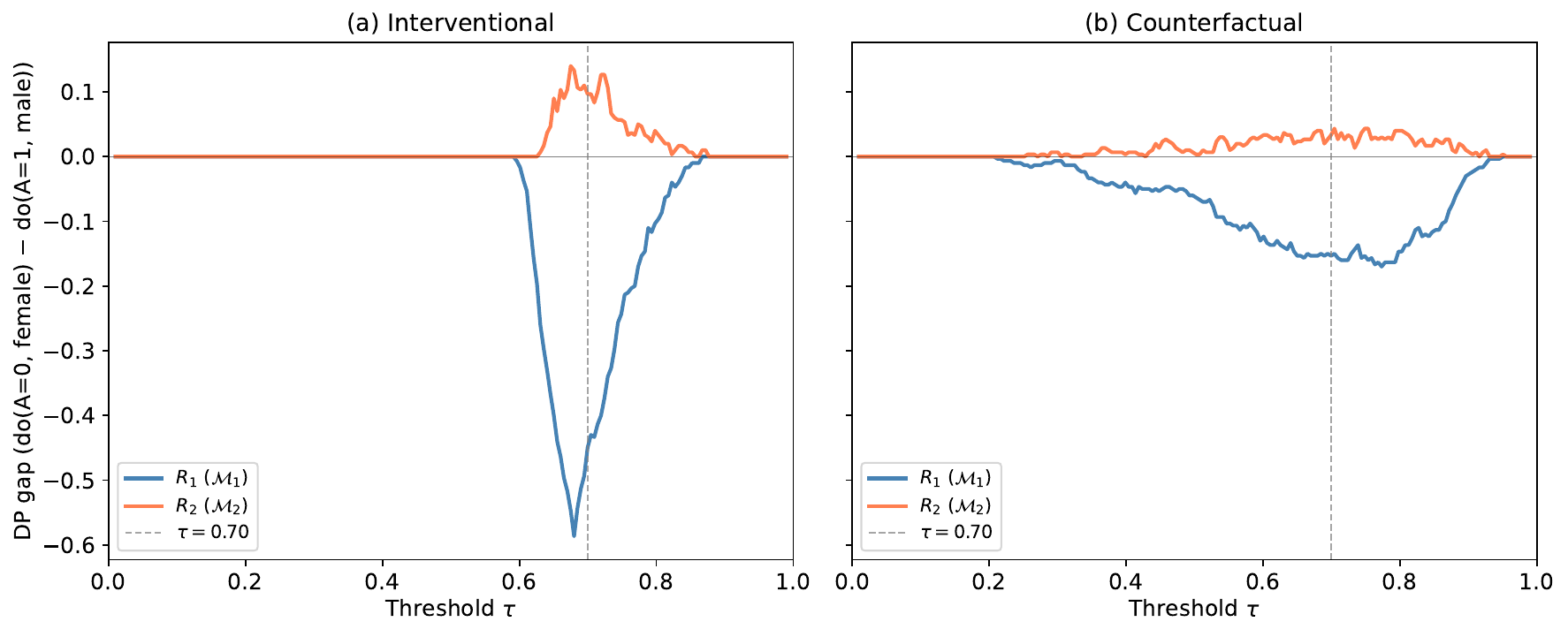}
\caption{DP gap as a function of the decision threshold $\tau$ for the two receivers under structural causal perception, where $A{=}0$ is female and $A{=}1$ is male. (a)~Interventional: $P(\hat{Y} \geq \tau \mid \text{do}(A{=}0)) - P(\hat{Y} \geq \tau \mid \text{do}(A{=}1))$. (b)~Counterfactual: same quantity under abduction. The dashed line marks $\tau = 0.70$.}
\label{fig:causal_dp_sweep}
\end{figure}

Let us now turn to the \textit{interventional and counterfactual predicted distributions}. Rather than fixing a single threshold, we sweep $\tau$ from 0 to 1 and compute the causal DP gap at each point: $P(\hat{Y} \geq \tau \mid \text{do}(A{=}0)) - P(\hat{Y} \geq \tau \mid \text{do}(A{=}1))$, where $A{=}0$ denotes female and $A{=}1$ male. Fig.\@~\ref{fig:causal_dp_sweep} plots these curves for both receivers at the interventional (2nd rung) and counterfactual (3rd rung) levels. The pattern is consistent across all thresholds. Under $R_1$, the interventional DP gap is negative throughout the relevant range of $\tau$, peaking at $-0.450$ at $\tau = 0.70$: if everyone were set to female, far fewer applicants would be approved than if everyone were set to male. Under $R_2$, the gap is near zero or slightly positive ($+0.097$ at $\tau = 0.70$), indicating that removing the direct edge $A \to Y$ substantially reduces the causal gender effect on decisions in this setup. At the counterfactual level, the same divergence holds but less: $R_1$ shows a gap of $-0.150$ and $R_2$ of $+0.037$ at $\tau = 0.70$. The attenuation reflects the fact that counterfactual reasoning preserves individual-specific noise, partially absorbing the gender effect through observed mediators.
Table~\ref{tab:within_model} also reports within-model distances for the 3rd rung. These distances are population-level counterfactual sensitivity summaries, not individual-level tests of counterfactual fairness in the strict sense of Kusner et al.\ \cite{DBLP:conf/nips/KusnerLRS17}. We once again see a difference in fairness interpretation with $R_1$ showing larger distances than $R_2$. 
% The parametrical within-model pattern is analogous (the baseline $\mathcal{M}_1$ is identical; $\mathcal{M}_2$'s weaker $A \to Y$ coefficient yields comparably small within-model distances).

%

These competing fairness results stress the notion of situated bias (Section~\ref{sec:Background}). Under a fairness view that disfavors a direct $A \to Y$ path, one may prefer receiver $R_2$ over $R_1$.
% \footnote{Unsurprisingly given that $R_1$ allows for $A \rightarrow Y$; experiments are illustrative.} 
This preference is a normative judgment about the reference SCM, not an output of the causal perception metric itself. In Example~\ref{ex:CreditScoring}, even if both loan officers have equal standing, the framework lets us state explicitly that $R_1$ appears biased with respect to the reference worldview represented by $R_2$.

% ============================================================
\section{Discussion}
\label{sec:Discussion}
% ============================================================

We presented the first implementation of the causal perception framework \cite{Alvarez2025CausalPerception}. We designed two algorithms for single- and multi-intervention settings at both the interventional and counterfactual levels (the 2nd and 3rd rungs of Pearl's causal ladder, respectively), and we studied a set of distance functions. These allowed us to measure causal perception at different levels under competing SCMs. We showcased the proposed implementation on real data, motivating the need to consider causal perception as a phenomenon that can affect accurate and fair decision-making in multi-expert settings.

In practice, implementing causal perception requires choosing a distance measure, an aggregation function, and a threshold. Our results give practical guidance for these choices. The three distances answer different questions: $W_2$ measures how far apart the distributions sit on the probability scale, making it suitable when the magnitude of the disagreement matters; $D_{\mathrm{KL}}$ reacts to mismatches in density overlap, making it sensitive at the interventional level, though it attenuates under abduction and depends on the stability of the density estimate; and $\mathrm{TV}$ is bounded in $[0,1]$, giving the threshold $\epsilon$ a direct probability-scale reading. Because the perception verdict can flip across metrics for the same pair of SCMs (Section~\ref{sec:Experiments}), auditors should report all three distances with their confidence intervals rather than commit to one. The same holds for the threshold: sweeping $\epsilon$ exposes how robust a verdict is to the tolerance level. When a single system-level verdict is required, the aggregate $\bar{d}$ should be read alongside the per-intervention distances $\boldsymbol{\Delta}$ to locate where the disagreement concentrates.

\subsection{Limitations}

Three limitations are worth noting for the current implementation. 
First, we take the causal graph for granted and do not provide a method for learning it. Although we view causal perception as a consequence of competing causal worldviews, drawing the causal graph is a valid and contested concern when using SCMs in practice. The use of causal discovery algorithms \cite{Peters2017} or even LLMs \cite{DBLP:journals/tmlr/KicimanN0T24} is an interesting direction for future work.

Second, we work with ANMs to parametrize the structural equations. This is a strong functional assumption on the SCM that may not hold in all cases. Alternative parametrizations, such as causal normalizing flows \cite{DBLP:conf/nips/JavaloySV23}, are worth exploring. It remains an open question whether parametrical causal perception materializes under more lax functional forms for the causal relationships.

Third, we explore a case of competing SCMs. Our baseline is Fig.\@~\ref{fig:chiappa_dag} of Chiappa \cite{DBLP:conf/aaai/Chiappa19}, which is one of many possible SCMs for the German Credit dataset. From it, we present an alternative SCM for structural and parametrical causal perception based on the edge $A \rightarrow Y$. These alternatives are straightforward for illustrative purposes. The experimental results remain limited by our choice of competing SCMs. The illustrative nature of the experiments is intentional: we choose a well-understood manipulation precisely to validate the implementation against known intuitions before applying it to more complex disagreements.

\subsection{Future Work}

As the state of the art aims for agentic systems, it remains to be seen whether causal perception emerges within such a context. This direction would require extending the implementation of causal perception and, potentially, revisiting the original formulation by \'Alvarez and Ruggieri \cite{Alvarez2025CausalPerception}. It will also align with that work's proposed theory on how and why agents construct competing SCMs through the processes of signification and categorization.

Recent work \cite{DBLP:conf/nips/JinCLGKLBAKSS23,DBLP:journals/tmlr/KicimanN0T24,DBLP:conf/clear2/ChatziBSTG25} already explores whether LLMs can reason causally in the Pearlian sense. A natural next step would be to link this line of work with causal perception, exploring the possibility that different models reason differently in a causal sense about the same information. This too extends to agentic systems, where we may want to explore if agents, each equipped with its own SCM, can exhibit causal perception when interacting with each other or if causal perception itself is an emergent property of causal agents. 

Beyond this line of work, more complex implementations should be explored, including efficient algorithms and the problem of exploring a large space of possible interventions. Future work should also explore reconciliation strategies for multi-expert settings where causal perception is detected.

% ============================================================
% Credits
% ============================================================

\begin{credits}
\subsubsection{\discintname}
The author has no competing interests to declare that are relevant to this paper.
The paper reflects only the author’s views.
The Santander AI Lab is not responsible for any use that may be made of the information it contains.
\end{credits}

% ---- Bibliography ----
\bibliographystyle{splncs04}
\bibliography{references}

@inproceedings{Alvarez2025CausalPerception,
  title={Toward A Causal Framework for Modeling Perception},
  author = {Jos{\'{e}} M. {\'{A}}lvarez and Salvatore Ruggieri},
  booktitle={Proceedings of the AAAI/ACM Conference on AI, Ethics, and Society},
  volume={8},
  pages={166--178},
  year={2025}
}

@book{PearlCausality2009,
    author = {Pearl, Judea},
    edition = {2nd},
    publisher = {Cambridge University Press},
    title = {Causality: Models, Reasoning, and Inference},
    year = 2009
}

@book{Pearl2018WHy,
    author = {Pearl, Judea and Mackenzie, Dana},
    publisher = {Basic Books},
    title = {The Book of Why: {T}he New Science of Cause and Effect},
    year = 2018
}

@inproceedings{DBLP:conf/nips/KusnerLRS17,
  author       = {Matt J. Kusner and
                  Joshua R. Loftus and
                  Chris Russell and
                  Ricardo Silva},
  title        = {Counterfactual Fairness},
  booktitle    = {{NIPS}},
  pages        = {4066--4076},
  year         = {2017}
}

@inproceedings{DBLP:conf/nips/RussellKLS17,
  author       = {Chris Russell and
                  Matt J. Kusner and
                  Joshua R. Loftus and
                  Ricardo Silva},
  title        = {When Worlds Collide: Integrating Different Counterfactual Assumptions
                  in Fairness},
  booktitle    = {{NIPS}},
  pages        = {6414--6423},
  year         = {2017}
}

@inproceedings{DBLP:conf/aaai/Chiappa19,
  author       = {Silvia Chiappa},
  title        = {Path-Specific Counterfactual Fairness},
  booktitle    = {{AAAI}},
  pages        = {7801--7808},
  publisher    = {{AAAI} Press},
  year         = {2019}
}

@inproceedings{DBLP:conf/fat/BaumannCCIR23,
  author       = {Joachim Baumann and
                  Alessandro Castelnovo and
                  Riccardo Crupi and
                  Nicole Inverardi and
                  Daniele Regoli},
  title        = {Bias on Demand: {A} Modelling Framework That Generates Synthetic Data
                  With Bias},
  booktitle    = {FAccT},
  pages        = {1002--1013},
  publisher    = {{ACM}},
  year         = {2023}
}

@inproceedings{DBLP:conf/nips/JavaloySV23,
  author       = {Adri{\'{a}}n Javaloy and
                  Pablo S{\'{a}}nchez{-}Mart{\'{\i}}n and
                  Isabel Valera},
  title        = {Causal normalizing flows: from theory to practice},
  booktitle    = {NeurIPS},
  year         = {2023}
}

@article{DBLP:journals/jair/AlvarezR25,
  author       = {Jos{\'{e}} M. {\'{A}}lvarez and
                  Salvatore Ruggieri},
  title        = {Counterfactual Situation Testing: From Single to Multidimensional
                  Discrimination},
  journal      = {J. Artif. Intell. Res.},
  volume       = {82},
  pages        = {2279--2323},
  year         = {2025}
}

@article{DBLP:journals/corr/abs-2405-13693,
    author = {Jos{\'{e}} M. {\'{A}}lvarez and Salvatore Ruggieri},
    title = {Mutatis Mutandis: Revisiting the Comparator in Discrimination Testing},
    journal = {Computational Intelligence},
    volume = {42},
    number = {4},
    pages = {e70238},
    year = {2026}
}

@misc{KahnemanNIPS2021,
    title = {NeurIPS 2021: {A} Conversation on Human and Machine Intelligence},
    author = {Kahneman, Daniel},
    year = {2021},
    howpublished = {\url{https://nips.cc/virtual/2021/invited-talk/22284}}
}

@inproceedings{DBLP:conf/uai/RubensteinWBMJG17,
  author       = {Paul K. Rubenstein and
                  Sebastian Weichwald and
                  Stephan Bongers and
                  Joris M. Mooij and
                  Dominik Janzing and
                  Moritz Grosse{-}Wentrup and
                  Bernhard Sch{\"{o}}lkopf},
  title        = {Causal Consistency of Structural Equation Models},
  booktitle    = {{UAI}},
  publisher    = {{AUAI} Press},
  year         = {2017}
}

@book{Villani2009OptimalTransport,
  author    = {C{\'{e}}dric Villani},
  title     = {Optimal Transport: Old and New},
  publisher = {Springer},
  year      = {2009}
}

@article{KullbackLeibler1951,
  author  = {Solomon Kullback and Richard A. Leibler},
  title   = {On Information and Sufficiency},
  journal = {The Annals of Mathematical Statistics},
  volume  = {22},
  number  = {1},
  pages   = {79--86},
  year    = {1951}
}

@book{Tsybakov2009NonparametricEstimation,
  author    = {Alexandre B. Tsybakov},
  title     = {Introduction to Nonparametric Estimation},
  publisher = {Springer},
  year      = {2009}
}

@article{tversky_judgment_1974,
	title = {Judgment {Under} {Uncertainty}: {Heuristics} and {Biases}},
	volume = {185},
	number = {4157},
	journal = {Science},
	author = {Tversky, Amos and Kahneman, Daniel},
	year = {1974},
	pages = {1124--1131},
}

@inproceedings{DBLP:conf/kdd/SrivastavaHK19,
  author       = {Megha Srivastava and
                  Hoda Heidari and
                  Andreas Krause},
  title        = {Mathematical Notions vs. Human Perception of Fairness: {A} Descriptive
                  Approach to Fairness for Machine Learning},
  booktitle    = {{KDD}},
  pages        = {2459--2468},
  publisher    = {{ACM}},
  year         = {2019}
}

@inproceedings{DBLP:conf/aies/Yaghini0H21,
  author       = {Mohammad Yaghini and
                  Andreas Krause and
                  Hoda Heidari},
  title        = {A Human-in-the-loop Framework to Construct Context-aware Mathematical
                  Notions of Outcome Fairness},
  booktitle    = {{AIES}},
  pages        = {1023--1033},
  publisher    = {{ACM}},
  year         = {2021}
}

@inproceedings{DBLP:conf/aies/BertrandBEM22,
  author       = {Astrid Bertrand and
                  Rafik Belloum and
                  James R. Eagan and
                  Winston Maxwell},
  title        = {How Cognitive Biases Affect XAI-assisted Decision-making: {A} Systematic
                  Review},
  booktitle    = {{AIES}},
  pages        = {78--91},
  publisher    = {{ACM}},
  year         = {2022}
}

@inproceedings{DBLP:conf/icdm/CaldersKP09,
  author       = {Toon Calders and
                  Faisal Kamiran and
                  Mykola Pechenizkiy},
  title        = {Building Classifiers with Independency Constraints},
  booktitle    = {{ICDM} Workshops},
  pages        = {13--18},
  publisher    = {{IEEE} Computer Society},
  year         = {2009}
}

@inproceedings{DBLP:conf/nips/HardtPNS16,
  author       = {Moritz Hardt and
                  Eric Price and
                  Nati Srebro},
  title        = {Equality of Opportunity in Supervised Learning},
  booktitle    = {{NIPS}},
  pages        = {3315--3323},
  year         = {2016}
}

@article{DBLP:journals/tmlr/KicimanN0T24,
  author       = {Emre Kiciman and
                  Robert Osazuwa Ness and
                  Amit Sharma and
                  Chenhao Tan},
  title        = {Causal Reasoning and Large Language Models: Opening a New Frontier
                  for Causality},
  journal      = {Trans. Mach. Learn. Res.},
  volume       = {2024},
  year         = {2024}
}

@inproceedings{DBLP:conf/clear2/ChatziBSTG25,
  author       = {Ivi Chatzi and
                  Nina L. Corvelo Benz and
                  Eleni Straitouri and
                  Stratis Tsirtsis and
                  Manuel Gomez{-}Rodriguez},
  title        = {Counterfactual Token Generation in Large Language Models},
  booktitle    = {CLeaR},
  series       = {Proceedings of Machine Learning Research},
  pages        = {1291--1315},
  publisher    = {{PMLR}},
  year         = {2025}
}

@inproceedings{DBLP:conf/aaai/Ruggieri0PST23,
  author       = {Salvatore Ruggieri and
                  Jos{\'{e}} M. {\'{A}}lvarez and
                  Andrea Pugnana and
                  Laura State and
                  Franco Turini},
  title        = {Can We Trust Fair-{AI}?},
  booktitle    = {{AAAI}},
  pages        = {15421--15430},
  publisher    = {{AAAI} Press},
  year         = {2023}
}

@inproceedings{DBLP:conf/icse/VermaR18,
  author       = {Sahil Verma and
                  Julia Rubin},
  title        = {Fairness definitions explained},
  booktitle    = {FairWare@ICSE},
  pages        = {1--7},
  publisher    = {{ACM}},
  year         = {2018}
}

@article{DBLP:journals/kais/GoldenbergW19,
  author       = {Igor Goldenberg and
                  Geoffrey I. Webb},
  title        = {Survey of distance measures for quantifying concept drift and shift
                  in numeric data},
  journal      = {Knowl. Inf. Syst.},
  volume       = {60},
  number       = {2},
  pages        = {591--615},
  year         = {2019}
}

@book{Peters2017,
    author = {Jonas Peters and Dominik Janzing and Bernhard Sch{\"{o}}lkopf},
    publisher = {MIT Press},
    title = {Elements of Causal Inference: Foundations and Learning Algorithms},
    year = 2017
}

@article{DBLP:journals/corr/abs-2601-03203,
  author       = {Davi Val{\'{e}}rio and
                  Chrysoula Zerva and
                  Mariana Pinto and
                  Ricardo Santos and
                  Andr{\'{e}} V. Carreiro},
  title        = {Counterfactual Fairness with Graph Uncertainty},
  journal      = {CoRR},
  volume       = {abs/2601.03203},
  year         = {2026}
}

@inproceedings{DBLP:conf/uai/KilbertusBKWS19,
  author    = {Niki Kilbertus and
               Philip J. Ball and
               Matt J. Kusner and
               Adrian Weller and
               Ricardo Silva},
  title     = {The Sensitivity of Counterfactual Fairness to Unmeasured Confounding},
  booktitle = {{UAI}},
  series    = {Proceedings of Machine Learning Research},
  volume    = {115},
  pages     = {616--626},
  publisher = {{AUAI} Press},
  year      = {2019}
}

@inproceedings{DBLP:conf/nips/JinCLGKLBAKSS23,
  author       = {Zhijing Jin and
                  Yuen Chen and
                  Felix Leeb and
                  Luigi Gresele and
                  Ojasv Kamal and
                  Zhiheng Lyu and
                  Kevin Blin and
                  Fernando Gonzalez Adauto and
                  Max Kleiman{-}Weiner and
                  Mrinmaya Sachan and
                  Bernhard Sch{\"{o}}lkopf},
  title        = {CLadder: {A} Benchmark to Assess Causal Reasoning Capabilities of
                  Language Models},
  booktitle    = {NeurIPS},
  year         = {2023}
}

@inproceedings{DBLP:journals/corr/abs-2502-01211,
  author = {Bothmann, Ludwig and Boustani, Philip A. and Alvarez, Jose M. and Casalicchio, Giuseppe and Bischl, Bernd and Dandl, Susanne},
  title = {Privilege Scores},
  year = {2026},
  booktitle = {FAccT},
  publisher    = {{ACM}},
  pages = {4293--4341}
}

@article{DBLP:journals/statsci/Breiman01,
  author    = {Leo Breiman},
  title     = {Statistical Modeling: The Two Cultures},
  journal   = {Statistical Science},
  volume    = {16},
  number    = {3},
  pages     = {199--231},
  year      = {2001}
}

@inproceedings{DBLP:conf/icml/MarxCU20,
  author    = {Charles T. Marx and
               Flavio P. Calmon and
               Berk Ustun},
  title     = {Predictive Multiplicity in Classification},
  booktitle = {{ICML}},
  series    = {Proceedings of Machine Learning Research},
  volume    = {119},
  pages     = {6765--6774},
  publisher = {{PMLR}},
  year      = {2020}
}

\newpage
\appendix

% ============================================================
\section{Supplementary Material}
\label{app:Supp_Material}
% ============================================================

Let $A$ denote the protected attribute and $Y$ the outcome, with $Y=1$ denoting a positive outcome like being approved for a loan. Let $\hat{Y}$, in turn, denote the predicted outcome or decision.\footnote{We can also treat it as a predicted probability that we then discretize using a threshold $\tau \in \mathbb{R}^+$.} Consider a receiver $R_i$ with SCM $\mathcal{M}_{R_i}$ and its implied $\mathcal{P}_{\mathbf{X}_{R_i}}$. For our purposes, $A \in \{a', a\}$ is binary and $R_i$ constructs $\hat{Y}$. 

\paragraph{Demographic parity} (DP) \cite{DBLP:conf/icdm/CaldersKP09} requires that $\hat{Y}$ is independent of $A$:
\begin{equation}
\label{eq:DP}
    P_{R_i}(\hat{Y} = 1 \mid A = a') = P_{R_i}(\hat{Y} = 1 \mid A = a).
\end{equation}
The DP gap, in turn, measures the violation of demographic parity: $\Delta_{\mathrm{DP}} = P_{R_i}(\hat{Y}{=}1 \mid A{=}a') - P_{R_i}(\hat{Y}{=}1 \mid A{=}a)$.

\paragraph{Equalized odds} (EO) \cite{DBLP:conf/nips/HardtPNS16} requires that $\hat{Y}$ is independent of $A$ conditional on $Y$:
\begin{equation}
\label{eq:EO}
    P_{R_i}(\hat{Y} = 1 \mid Y = y, A = a') = P_{R_i}(\hat{Y} = 1 \mid Y = y, A = a), \quad \forall\, y \in \{0, 1\}.
\end{equation}
This decomposes, in practice, into a true positive rate (TPR) gap at $Y=1$ and a false positive rate (FPR) gap at $Y=0$.

\paragraph{Counterfactual fairness} \cite{DBLP:conf/nips/KusnerLRS17} requires that $R_i$'s prediction would remain the same had the individual belonged to a different group, given $R_i$'s SCM:
\begin{equation}
\label{eq:CF}
    P_{R_i}(\hat{Y}_{A \leftarrow a}(\mathbf{U}) \mid \mathbf{X} = \mathbf{x}, A = a) = P_{R_i}(\hat{Y}_{A \leftarrow a'}(\mathbf{U}) \mid \mathbf{X} = \mathbf{x}, A = a).
\end{equation}

All three fairness criteria are situated with respect to $R_i$. They depend on the decisions (or predictions) produced by a specific receiver's SCM. Two receivers with different SCMs may satisfy or violate the same criterion to different degrees, which is precisely what causal perception captures in the fairness domain. This point is illustrated in Section~\ref{sec:Experiments}.

% ============================================================
\section{Additional Experiments}
\label{app:Add_Material}
% ============================================================

\subsection{Parametrical Causal Perception}

Tables~\ref{tab:parametrical_plus1SE_2nd} to \ref{tab:parametrical_plus2SE_3rd} report parametrical perception distances when $\mathcal{M}_2$ uses $\hat{\beta} + 1\mathrm{SE}$ and $\hat{\beta} + 2\mathrm{SE}$ for the $A \to Y$ coefficient, respectively. The main text uses $\hat{\beta} - 2\mathrm{SE}$. As in the main text, distances under $\mathrm{do}(A{=}0)$ are exactly zero for all parametrical variants because setting $A{=}0$ nullifies the perturbed coefficient; these rows are omitted from the tables.

\subsection{Alternative Structural Disagreement: Age}
\label{app:structural_age}

The main text considers structural disagreement on the edge $A \to Y$ (gender $\to$ credit risk). Here, we test whether causal perception arises when receivers disagree on the edge $C \to Y$ (age $\to$ credit risk), a non-protected variable. Receiver $R_1$ holds $\mathcal{M}_1$ (the full DAG in Fig.\@~\ref{fig:chiappa_dag}, which includes $C \to Y$), and $R_2$ holds $\mathcal{M}_2$ (identical except that $C \to Y$ is removed: age affects $Y$ only indirectly through mediators $S$ and $R$). We intervene on age using $\mathrm{do}(C{=}26)$ (P25, ``young'') and $\mathrm{do}(C{=}43)$ (P75, ``old'').

\paragraph{Interventional distributions.} Table~\ref{tab:structural_age_2ndRung} reports the distances. The pattern is qualitatively similar to the $A \to Y$ case: $W_2$ remains below $\epsilon = 0.1$ ($\bar{d} = 0.037$), while $D_{\mathrm{KL}}$ and $TV$ exceed the threshold by large margins. The within-model distances confirm that removing $C \to Y$ nearly eliminates the direct age effect ($W_2 = 0.002$ for $\mathcal{M}_2$ vs.\ $W_2 = 0.074$ for $\mathcal{M}_1$). The interventional distributions are very narrow (std $\approx 0.03$), producing $TV = 1.000$ because the histogram bins do not overlap despite close support ranges.

\paragraph{Counterfactual distributions.} Table~\ref{tab:structural_age_3rdRung} reports the counterfactual distances. The wider counterfactual distributions (std $\approx 0.14$) yield more informative $TV$ values. The $W_2$ distances ($\bar{d} = 0.035$) are comparable to the $A \to Y$ case ($\bar{d} = 0.037$), and the within-model pattern is preserved: $\mathcal{M}_1$ shows a substantial age effect ($W_2 = 0.071$) while $\mathcal{M}_2$'s is negligible ($W_2 = 0.002$). These results confirm that causal perception is not specific to the protected attribute $A$; it arises whenever the receivers disagree on a structurally consequential edge.

\subsection{Nonlinear Robustness Check}
\label{app:nonlinear}

The main-text experiments model $Y$ with logistic regression, which is linear in the log-odds. To test whether the perception pattern is an artifact of this linearity, we replace the outcome model with a gradient boosting model (GBM), implemented as a \texttt{GradientBoostingClassifier} (100 trees, depth 3, learning rate 0.1), while keeping the mediator equations linear (OLS). The same two structural DAGs ($\mathcal{M}_1$: full, $\mathcal{M}_2$: no $A \to Y$) and the same German Credit data are used.

\paragraph{Interventional distributions.} Table~\ref{tab:nonlinear_2ndRung} reports the distances. The aggregated $W_2 = 0.041$ is nearly identical to the logistic baseline ($W_2 = 0.039$); $D_{\mathrm{KL}}$ and $TV$ again exceed $\epsilon = 0.1$. The qualitative verdict is unchanged: causal perception is detected by $D_{\mathrm{KL}}$ and $TV$ but not by $W_2$. A notable difference is the within-model pattern: with GBM, $\mathcal{M}_2$ exhibits a \emph{larger} within-model gender effect ($W_2 = 0.111$) than $\mathcal{M}_1$ ($W_2 = 0.077$). This reversal occurs because GBM captures nonlinear interactions through mediators that partially recover the gender signal even when the direct $A \to Y$ edge is absent.

\paragraph{Counterfactual distributions.} Table~\ref{tab:nonlinear_3rdRung} reports the counterfactual distances. The between model $W_2$ ($\bar{d} = 0.018$) is roughly half the logistic value ($\bar{d} = 0.037$), and $D_{\mathrm{KL}} \approx 0$. Thus, neither $W_2$ nor $D_{\mathrm{KL}}$ detects causal perception at the counterfactual level under GBM. However, $TV$ ($\bar{d} = 0.185$) still exceeds $\epsilon = 0.1$, so causal perception is detected by at least one metric. The within-model distances are also attenuated: with GBM, $\mathcal{M}_1$ and $\mathcal{M}_2$ produce nearly identical within-model gender effects ($W_2 \approx 0.02$). Overall, the nonlinear outcome model dampens counterfactual differences but does not eliminate them. The metric-dependent verdict underscores a practical lesson: whether causal perception is flagged depends on the outcome model, the choice of distance metric, and the perception threshold.

\begin{table}[htp]
\centering
\caption{Parametrical perception (interventional) with $\hat{\beta}_{A \to Y} + 1\mathrm{SE} = 0.573$. Distances under $\mathrm{do}(A{=}0)$ are zero and omitted. Bracketed values are 95\% bootstrap confidence intervals.}
\label{tab:parametrical_plus1SE_2nd}
\resizebox{\textwidth}{!}{%
\begin{tabular}{llccc}
\toprule
Comparison & Intervention & $W_2$ & $D_{\mathrm{KL}}$ & TV \\
\midrule
$\mathcal{M}_1$ vs.\ $\mathcal{M}_2$ & $\mathrm{do}(A{=}1)$ & 0.035\;{\scriptsize[.035,.036]} & 0.787\;{\scriptsize[.659,.988]} & 0.420\;{\scriptsize[.410,.517]} \\
\midrule
Within $\mathcal{M}_1$ & $\mathrm{do}(A{=}0)$ vs.\ $\mathrm{do}(A{=}1)$ & 0.064\;{\scriptsize[.063,.065]} & 2.734\;{\scriptsize[2.329,3.350]} & 0.583\;{\scriptsize[.563,.683]} \\
Within $\mathcal{M}_2$ & $\mathrm{do}(A{=}0)$ vs.\ $\mathrm{do}(A{=}1)$ & 0.099\;{\scriptsize[.098,.100]} & 10.146\;{\scriptsize[8.763,11.832]} & 0.773\;{\scriptsize[.733,.823]} \\
\bottomrule
\end{tabular}}%
\end{table}
\begin{table}[htp]
\centering
\caption{Parametrical perception (counterfactual) with $\hat{\beta}_{A \to Y} + 1\mathrm{SE} = 0.573$. Distances under $\mathrm{do}(A{=}0)$ are zero and omitted. Bracketed values are 95\% bootstrap confidence intervals.}
\label{tab:parametrical_plus1SE_3rd}
\resizebox{\textwidth}{!}{%
\begin{tabular}{llccc}
\toprule
Comparison & Intervention & $W_2$ & $D_{\mathrm{KL}}$ & TV \\
\midrule
$\mathcal{M}_1$ vs.\ $\mathcal{M}_2$ & $\mathrm{do}(A{=}1)$ & 0.035\;{\scriptsize[.034,.036]} & 0.034\;{\scriptsize[.031,.048]} & 0.183\;{\scriptsize[.223,.353]} \\
\midrule
Within $\mathcal{M}_1$ & $\mathrm{do}(A{=}0)$ vs.\ $\mathrm{do}(A{=}1)$ & 0.061\;{\scriptsize[.060,.063]} & 0.088\;{\scriptsize[.081,.123]} & 0.207\;{\scriptsize[.247,.360]} \\
Within $\mathcal{M}_2$ & $\mathrm{do}(A{=}0)$ vs.\ $\mathrm{do}(A{=}1)$ & 0.096\;{\scriptsize[.094,.098]} & 0.221\;{\scriptsize[.206,.298]} & 0.280\;{\scriptsize[.307,.403]} \\
\bottomrule
\end{tabular}}%
\end{table}
\begin{table}[htp]
\centering
\caption{Parametrical perception (interventional) with $\hat{\beta}_{A \to Y} + 2\mathrm{SE} = 0.763$. Distances under $\mathrm{do}(A{=}0)$ are zero and omitted. Bracketed values are 95\% bootstrap confidence intervals.}
\label{tab:parametrical_plus2SE_2nd}
\resizebox{\textwidth}{!}{%
\begin{tabular}{llccc}
\toprule
Comparison & Intervention & $W_2$ & $D_{\mathrm{KL}}$ & TV \\
\midrule
$\mathcal{M}_1$ vs.\ $\mathcal{M}_2$ & $\mathrm{do}(A{=}1)$ & 0.067\;{\scriptsize[.066,.068]} & 5.349\;{\scriptsize[4.579,6.569]} & 0.653\;{\scriptsize[.617,.723]} \\
\midrule
Within $\mathcal{M}_1$ & $\mathrm{do}(A{=}0)$ vs.\ $\mathrm{do}(A{=}1)$ & 0.064\;{\scriptsize[.063,.065]} & 2.734\;{\scriptsize[2.309,3.414]} & 0.583\;{\scriptsize[.560,.677]} \\
Within $\mathcal{M}_2$ & $\mathrm{do}(A{=}0)$ vs.\ $\mathrm{do}(A{=}1)$ & 0.131\;{\scriptsize[.129,.132]} & 17.021\;{\scriptsize[15.498,18.585]} & 0.877\;{\scriptsize[.843,.917]} \\
\bottomrule
\end{tabular}}%
\end{table}
\begin{table}[htp]
\centering
\caption{Parametrical perception (counterfactual) with $\hat{\beta}_{A \to Y} + 2\mathrm{SE} = 0.763$. Distances under $\mathrm{do}(A{=}0)$ are zero and omitted. Bracketed values are 95\% bootstrap confidence intervals.}
\label{tab:parametrical_plus2SE_3rd}
\resizebox{\textwidth}{!}{%
\begin{tabular}{llccc}
\toprule
Comparison & Intervention & $W_2$ & $D_{\mathrm{KL}}$ & TV \\
\midrule
$\mathcal{M}_1$ vs.\ $\mathcal{M}_2$ & $\mathrm{do}(A{=}1)$ & 0.067\;{\scriptsize[.065,.069]} & 0.133\;{\scriptsize[.122,.192]} & 0.287\;{\scriptsize[.280,.393]} \\
\midrule
Within $\mathcal{M}_1$ & $\mathrm{do}(A{=}0)$ vs.\ $\mathrm{do}(A{=}1)$ & 0.061\;{\scriptsize[.060,.063]} & 0.088\;{\scriptsize[.079,.122]} & 0.207\;{\scriptsize[.250,.360]} \\
Within $\mathcal{M}_2$ & $\mathrm{do}(A{=}0)$ vs.\ $\mathrm{do}(A{=}1)$ & 0.128\;{\scriptsize[.125,.132]} & 0.426\;{\scriptsize[.395,.552]} & 0.373\;{\scriptsize[.377,.477]} \\
\bottomrule
\end{tabular}}%
\end{table}
\begin{table}[htp]
\centering
\caption{Alternative structural perception ($C \to Y$): interventional distances under $\mathrm{do}(C{=}26)$ (young) and $\mathrm{do}(C{=}43)$ (old). Bracketed values are 95\% bootstrap confidence intervals.}
\label{tab:structural_age_2ndRung}
\resizebox{\textwidth}{!}{%
\begin{tabular}{llccc}
\toprule
Comparison & Intervention & $W_2$ & $D_{\mathrm{KL}}$ & TV \\
\midrule
$\mathcal{M}_1$ vs.\ $\mathcal{M}_2$ & $\mathrm{do}(C{=}26)$ & 0.038\;{\scriptsize[.037,.038]} & 4.553\;{\scriptsize[4.021,5.292]} & 1.000\;{\scriptsize[1.000,1.000]} \\
$\mathcal{M}_1$ vs.\ $\mathcal{M}_2$ & $\mathrm{do}(C{=}43)$ & 0.036\;{\scriptsize[.035,.037]} & 2.873\;{\scriptsize[2.777,3.096]} & 1.000\;{\scriptsize[1.000,1.000]} \\
$\mathcal{M}_1$ vs.\ $\mathcal{M}_2$ & $\bar{d}$ (mean) & 0.037\;{\scriptsize[.036,.037]} & 3.713\;{\scriptsize[3.399,4.194]} & 1.000\;{\scriptsize[1.000,1.000]} \\
\midrule
Within $\mathcal{M}_1$ & $\mathrm{do}(C{=}26)$ vs.\ $\mathrm{do}(C{=}43)$ & 0.074\;{\scriptsize[.073,.074]} & 9.096\;{\scriptsize[8.087,10.249]} & 1.000\;{\scriptsize[1.000,1.000]} \\
Within $\mathcal{M}_2$ & $\mathrm{do}(C{=}26)$ vs.\ $\mathrm{do}(C{=}43)$ & 0.002\;{\scriptsize[.002,.002]} & 0.014\;{\scriptsize[.013,.016]} & 1.000\;{\scriptsize[1.000,1.000]} \\
\bottomrule
\end{tabular}}%
\end{table}
\begin{table}[htp]
\centering
\caption{Alternative structural perception ($C \to Y$): counterfactual distances under $\mathrm{do}(C{=}26)$ (young) and $\mathrm{do}(C{=}43)$ (old). Bracketed values are 95\% bootstrap confidence intervals.}
\label{tab:structural_age_3rdRung}
\resizebox{\textwidth}{!}{%
\begin{tabular}{llccc}
\toprule
Comparison & Intervention & $W_2$ & $D_{\mathrm{KL}}$ & TV \\
\midrule
$\mathcal{M}_1$ vs.\ $\mathcal{M}_2$ & $\mathrm{do}(C{=}26)$ & 0.037\;{\scriptsize[.035,.038]} & 0.043\;{\scriptsize[.038,.059]} & 0.223\;{\scriptsize[.237,.357]} \\
$\mathcal{M}_1$ vs.\ $\mathcal{M}_2$ & $\mathrm{do}(C{=}43)$ & 0.033\;{\scriptsize[.031,.035]} & 0.029\;{\scriptsize[.023,.050]} & 0.183\;{\scriptsize[.213,.343]} \\
$\mathcal{M}_1$ vs.\ $\mathcal{M}_2$ & $\bar{d}$ (mean) & 0.035\;{\scriptsize[.033,.036]} & 0.036\;{\scriptsize[.030,.055]} & 0.203\;{\scriptsize[.225,.350]} \\
\midrule
Within $\mathcal{M}_1$ & $\mathrm{do}(C{=}26)$ vs.\ $\mathrm{do}(C{=}43)$ & 0.071\;{\scriptsize[.069,.073]} & 0.154\;{\scriptsize[.136,.208]} & 0.293\;{\scriptsize[.297,.423]} \\
Within $\mathcal{M}_2$ & $\mathrm{do}(C{=}26)$ vs.\ $\mathrm{do}(C{=}43)$ & 0.002\;{\scriptsize[.002,.002]} & 0.000\;{\scriptsize[.000,.001]} & 0.057\;{\scriptsize[.060,.140]} \\
\bottomrule
\end{tabular}}%
\end{table}
\begin{table}[htp]
\centering
\caption{Nonlinear robustness check (GBM for $Y$): interventional distances under $\mathrm{do}(A{=}0)$ (female) and $\mathrm{do}(A{=}1)$ (male). Bracketed values are 95\% bootstrap confidence intervals.}
\label{tab:nonlinear_2ndRung}
\resizebox{\textwidth}{!}{%
\begin{tabular}{llccc}
\toprule
Comparison & Intervention & $W_2$ & $D_{\mathrm{KL}}$ & TV \\
\midrule
$\mathcal{M}_1$ vs.\ $\mathcal{M}_2$ & $\mathrm{do}(A{=}0)$ & 0.052\;{\scriptsize[.049,.056]} & 0.582\;{\scriptsize[.458,.776]} & 0.837\;{\scriptsize[.797,.873]} \\
$\mathcal{M}_1$ vs.\ $\mathcal{M}_2$ & $\mathrm{do}(A{=}1)$ & 0.030\;{\scriptsize[.027,.033]} & 0.194\;{\scriptsize[.169,.239]} & 0.373\;{\scriptsize[.323,.447]} \\
$\mathcal{M}_1$ vs.\ $\mathcal{M}_2$ & $\bar{d}$ (mean) & 0.041\;{\scriptsize[.038,.044]} & 0.388\;{\scriptsize[.313,.507]} & 0.605\;{\scriptsize[.560,.660]} \\
\midrule
Within $\mathcal{M}_1$ & $\mathrm{do}(A{=}0)$ vs.\ $\mathrm{do}(A{=}1)$ & 0.077\;{\scriptsize[.070,.084]} & 0.983\;{\scriptsize[.696,1.450]} & 0.753\;{\scriptsize[.710,.813]} \\
Within $\mathcal{M}_2$ & $\mathrm{do}(A{=}0)$ vs.\ $\mathrm{do}(A{=}1)$ & 0.111\;{\scriptsize[.107,.116]} & 1.172\;{\scriptsize[1.006,1.463]} & 0.937\;{\scriptsize[.907,.963]} \\
\bottomrule
\end{tabular}}%
\end{table}
\begin{table}[htp]
\centering
\caption{Nonlinear robustness check (GBM for $Y$): counterfactual distances under $\mathrm{do}(A{=}0)$ (female) and $\mathrm{do}(A{=}1)$ (male). Bracketed values are 95\% bootstrap confidence intervals. Negative $D_{\mathrm{KL}}$ estimates, a KDE estimation artifact, are reported as 0.000.}
\label{tab:nonlinear_3rdRung}
\resizebox{\textwidth}{!}{%
\begin{tabular}{llccc}
\toprule
Comparison & Intervention & $W_2$ & $D_{\mathrm{KL}}$ & TV \\
\midrule
$\mathcal{M}_1$ vs.\ $\mathcal{M}_2$ & $\mathrm{do}(A{=}0)$ & 0.021\;{\scriptsize[.018,.029]} & 0.000\;{\scriptsize[.000,.004]} & 0.183\;{\scriptsize[.203,.327]} \\
$\mathcal{M}_1$ vs.\ $\mathcal{M}_2$ & $\mathrm{do}(A{=}1)$ & 0.014\;{\scriptsize[.012,.021]} & 0.007\;{\scriptsize[.002,.019]} & 0.187\;{\scriptsize[.193,.293]} \\
$\mathcal{M}_1$ vs.\ $\mathcal{M}_2$ & $\bar{d}$ (mean) & 0.018\;{\scriptsize[.015,.025]} & 0.000\;{\scriptsize[.000,.012]} & 0.185\;{\scriptsize[.198,.310]} \\
\midrule
Within $\mathcal{M}_1$ & $\mathrm{do}(A{=}0)$ vs.\ $\mathrm{do}(A{=}1)$ & 0.024\;{\scriptsize[.017,.047]} & 0.000\;{\scriptsize[.000,.017]} & 0.250\;{\scriptsize[.247,.373]} \\
Within $\mathcal{M}_2$ & $\mathrm{do}(A{=}0)$ vs.\ $\mathrm{do}(A{=}1)$ & 0.021\;{\scriptsize[.017,.047]} & 0.001\;{\scriptsize[.000,.032]} & 0.183\;{\scriptsize[.207,.327]} \\
\bottomrule
\end{tabular}}%
\end{table}

\end{document}